%% file: naacl2021.tex
\pdfoutput=1

\documentclass[11pt]{article}

\usepackage{authblk}
\usepackage[]{naacl2021}

\usepackage{times}
\usepackage{latexsym}
\usepackage{todonotes}

\usepackage[T1]{fontenc}

\usepackage[utf8]{inputenc}

\usepackage{microtype}
\usepackage{makecell}
\usepackage{multirow}
\usepackage{booktabs}
\usepackage{diagbox}
\usepackage{fixltx2e}

\title{
{
\textsl{``I'm Not Mad''}}: \\
Commonsense Implications of  Negation and Contradiction
}

\author[1]{Liwei Jiang}
\author[2,3]{Antoine Bosselut}
\author[2]{Chandra Bhagavatula}
\author[1,2]{Yejin Choi}
\affil[1]{Paul G. Allen School of Computer Science \& Engineering, University of Washington} 
\affil[2]{Allen Institute for Artificial Intelligence}
\affil[3]{Stanford University}
\affil[ ]{\tt \{lwjiang,yejin\}@cs.washington.edu,\{chandrab\}@allenai.org}

\usepackage{xcolor}
\definecolor{colorxmark}{RGB}{255, 87, 51}
\definecolor{colorcmark}{RGB}{66, 154, 137}

\usepackage{color, colortbl}
\definecolor{Gray}{gray}{0.9}

\usepackage{amssymb}
\usepackage{pifont}
\newcommand{\cmark}{\color{colorcmark} \textbf{\ding{51}}}%
\newcommand{\xmark}{\color{colorxmark} \textbf{\ding{55}}}%

\newcommand{\full}{\textsc{Full}}

\newcommand{\dataset}{\textsc{Anion}}
\newcommand{\atomic}{\textsc{Atomic}}

\newcommand{\logical}{logical}
\newcommand{\Logical}{Logical}
\newcommand{\semilogical}{semi-logical}
\newcommand{\Semilogical}{Semi-logical}

\newcommand{\Contradiction}{Commonsense Contradiction}

\newcommand{\Patlenvalid}{Precision @ \{\# valid\}}
\newcommand{\patlenvalid}{P@\{\# valid\}}

\newcommand{\Patthree}{Precision @ 3}
\newcommand{\patthree}{P@3}

\newcommand{\Patten}{Precision @ 10}
\newcommand{\patten}{P@10}

\newcommand{\eg}{\textit{e.g.}}
\newcommand{\ie}{\textit{i.e.}}

\begin{document}
\maketitle
\begin{abstract}
Natural language inference requires reasoning about contradictions, negations, and their commonsense implications. 
Given a simple premise (e.g., ``I'm mad at you''), humans can reason about the varying shades of contradictory statements ranging from straightforward negations (``I'm not mad at you'') to commonsense contradictions (``I'm happy''). Moreover, these negated or contradictory statements shift the commonsense implications of the original premise in nontrivial ways. For example, while ``I'm mad'' implies ``I'm unhappy about something,'' negating the premise (\ie, ``I'm \textit{not} mad'') does not necessarily negate the corresponding commonsense implications. 

In this paper, we present the first comprehensive study focusing on commonsense implications of negated statements and contradictions. We introduce \dataset{}\footnote{Data and code available at \url{https://github.com/liweijiang/anion}}, a new commonsense knowledge graph with 624K \textit{if-then} rules focusing on negated and contradictory events. We then present joint generative and discriminative inference models for this new resource, providing novel empirical insights on how logical negations and commonsense contradictions reshape the commonsense implications of their original premises. 
\end{abstract}

\input{sections/1-intro}

\input{sections/2-negation_background}

\input{sections/3-resource}

\input{sections/4-comet-neg}

\input{sections/5-discriminator}

\input{sections/6-discussion}


\input{sections/8-conclusion}

\input{sections/ethical}

\section*{Acknowledgements}

The authors thank Elisa Kreiss for helpful discussions. We also thank the anonymous reviewers and meta-reviewers for their helpful feedback.
This research was supported in part by DARPA under the MCS program through NIWC Pacific (N66001-19-2-4031), and the Allen Institute for AI (AI2).


\bibliography{anthology,custom}
\bibliographystyle{acl_natbib}

\newpage
\include{sections/appendix}

\end{document}

%% file: sections/1-intro.tex
\section{Introduction}
\label{sec:intro}

\begin{figure}[t]
    \centering
    \includegraphics[width=0.48\textwidth]{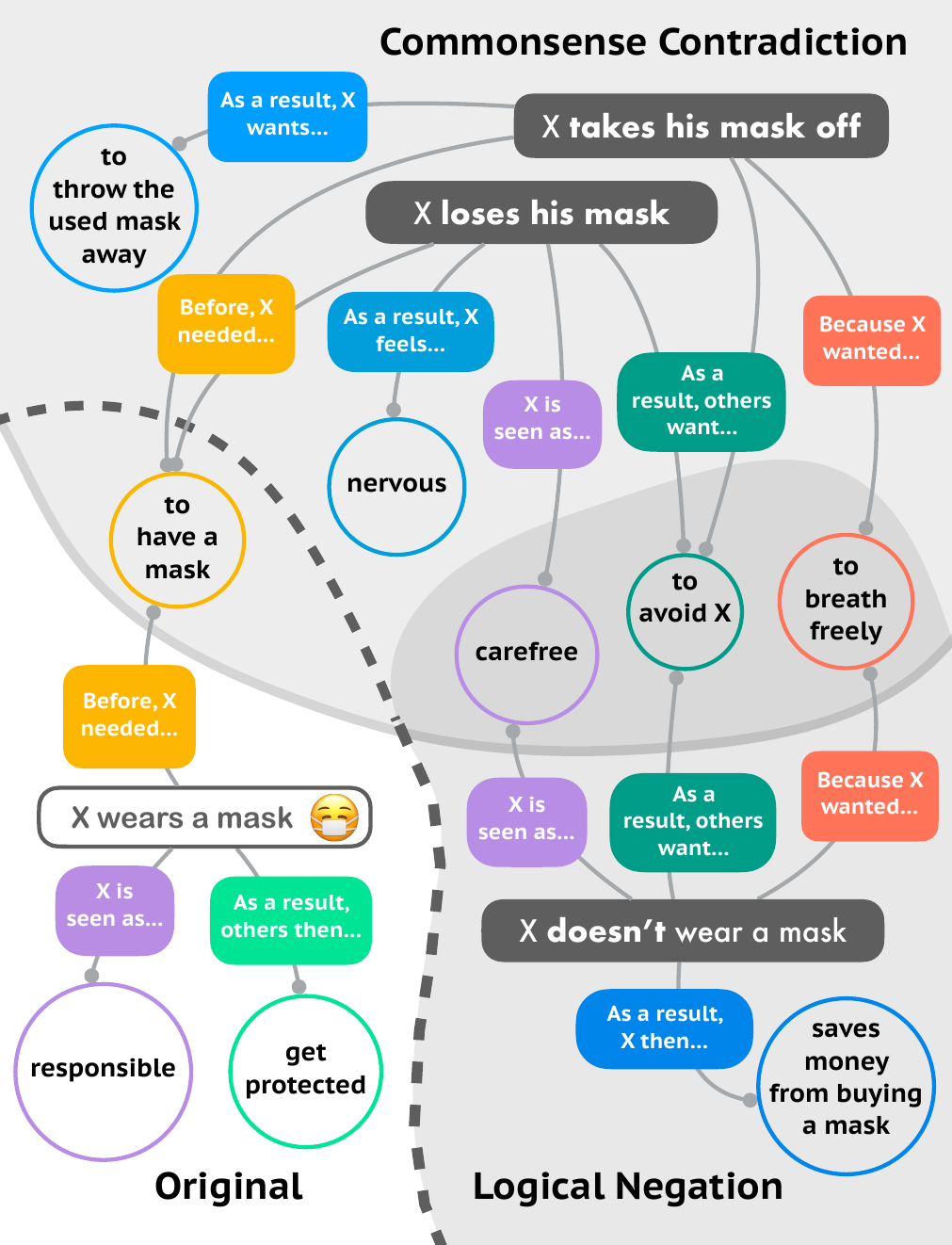}
    \caption{Commonsense inferences for the event ``X wears a mask,'' its logical negation and commonsense contradiction events, and their associated inferences.}
    \label{fig:example_pos_neg_inference}
\end{figure}

Humans reason about underlying causes and effects of events described in text. For example, in Figure \ref{fig:example_pos_neg_inference}, the event ``X wears a mask'' is associated with many causal inferences such as ``X is seen as responsible,'' or ``Others get protected.'' 
Hypothesizing and reasoning about commonsense inferences is used for understanding complex situations encountered in everyday life \cite{sap2019socialiqa, bisk2019piqa, bhagavatula2020abductive, Sakaguchi2020WINOGRANDEAA}. 
This ability eludes AI systems, and has motivated the design of a wealth of commonsense knowledge resources, 
such as Cyc \cite{Lenat1995Cyc}, 
ConceptNet \cite{speer2018conceptnet}, and 
\atomic{} \citep{sap2019atomic, hwang2020cometatomic}, 
 to provide structured reasoning capabilities to AI systems \cite{lin2019kagnet, feng2020scalable}.


However, reasoning about 
negated observations remains a challenge \cite{hossain2020its}. 
While negation is often considered a poorer form of meaning than affirmation\footnote{Following \citet{sep-negation}, we classify declarative expressions as affirmations or negations/contradictions based on whether they affirm or deny an action or object.} \cite{ackrill1963aristotle, sep-negation}, negated statements can still imply expressive commonsense inferences. In Figure~\ref{fig:example_pos_neg_inference}, the negated event “X doesn't wear a mask,” 
is connected to rich commonsense inferences, despite describing the \textit{absence} of action. However, negated observations are rarely found in commonsense knowledge resources. For example, negated examples make up only $\sim$3\% of examples in the ConceptNet knowledge graph \citep{li-etal-2016-commonsense}.

This scarcity poses downstream issues for systems that must understand negated situations. Commonsense knowledge models \citep{bosselut2019comet, hwang2020cometatomic} trained on resources of largely affirmative instances struggle particularly with negation examples. Their ability to hypothesize inferences for negated events is  35\% lower than for affirmative events (\S\ref{ssec:model-comet:exps}).
Furthermore, 
since negated statements are asymmetrically mentioned in text compared to affirmative statements \citep{jowett1892dialogues,sep-negation}, large-scale pretraining does not implicitly learn negation scoping \citep{kim-etal-2019-probing}. As a result, when presented with negated concepts, pretrained neural language models (PLMs) often exhibit the same associations as affirmative statements \cite{kassner-etal-2020-pretrained}. Motivated by these observations, our work focuses on 
improving the ability of knowledge models to make commonsense inferences about events that convey denial, rejection or contradiction of actions.

We define our contributions as follows. First, we crowdsource a new large scale resource, \textbf{A}rray of commonse\textbf{N}se \textbf{I}nferences for \textbf{O}ppositions and \textbf{N}egations (\dataset{}), which contains inferences for different types of negated events.
This new resource can be used to train knowledge models on commonsense inferences associated with the absence of actions. Second, we propose a new class of negation discriminators that can be applied to generated commonsense inferences. These discriminators partition inferences based on logical consistency, thereby mitigating the effects of common affirmative associations that violate negation constraints. Discriminators are trained using contrastive samples from paired affirmative and negated events in \dataset. Finally, we conduct an  empirical study of both of these techniques and show that using training- and discriminator-based approaches for modeling negation cuts the performance difference between affirmative and negated events by 73\% - 85\% depending on the negation variety.


%% file: sections/2-negation_background.tex
\section{Commonsense Negation}
\label{sec:negation_background}

\paragraph{Negation in Language}

In \textit{Categories} and \textit{De Interpretatione}, Aristotle classifies declarative statements into affirmation and negation, which respectively affirms or denies observations about an event \citep{ackrill1963aristotle}. 
Despite this seeming simplicity, natural language often expresses negation in complex and subtle ways, using diverse syntactic, semantic and pragmatic formulations 
\cite{sep-negation}. 
For example, syntactically, different negation determiners (\ie, negation cues) such as \textit{no}, \textit{few} and \textit{only} result in distinct explicit and implicit negative perceptions \cite{XIANG201671}.

Despite their diversity, however, negated language expressions are much less likely to appear in text than affirmative statements \cite{reitan-etal-2015-negation}. 
Consequently, PLMs, which rely on large-scale textual corpora as training data, are prone to decreased performance when confronted with negated constructions. 
In machine translation, for example, the presence of negation may heavily affect the quality of produced translations \cite{fancellu-webber-2015-translating-negation,hossain2020its}. 
In factual knowledge understanding tasks, PLMs memorize positive and negative sentences seen during training, but generalize more poorly to unseen negated instances \cite{kassner-schutze-2020-negated}.

\input{tables/atomic-neg-cue-example-table}

\paragraph{Negation in Commonsense Reasoning}

Understanding negation and oppositional expressions is critical for reasoning about commonsense knowledge, particularly in counterfactual scenarios \citep{qin-etal-2019-counterfactual}.  
However, negation is rarely explicitly modeled in NLP studies on commonsense reasoning. 
As a result, in many NLP tasks, these models experience a performance drop when presented with examples exhibiting negated characteristics.

As a case study, the \atomic{} \cite{sap2019atomic} knowledge graph encodes social commonsense knowledge about event pre-conditions, 
event post-conditions, 
and static attributes in the form of natural language \textit{if-then} rules.
However, despite the fact that \atomic{} provides a rich set of seed events, 
it comprises an unbalanced set of affirmative events (97.9\%) and negated events (2.1\%). 
As a result, when systems link to \atomic{} to retrieve relevant social commonsense inferences, they are likely to recover inferences of affirmative events even when searching for negated instances. 
Furthermore, knowledge models that use this resource 
(\eg, COMET; \citealp{bosselut2019comet}) are unlikely to learn implicit differences between inferences of affirmative and negated events.
When given negated events, these models often 
produce associations of counterpart affirmative events. 
For example, for the negated event, ``X opposes racism,'' COMET infers ``X intends to be a racist,'' an association of the affirmative statement, ``X supports racism.'' 

At the heart of this problem is that inferring commonsense knowledge about negations often requires implicit reasoning. 
In factual knowledge reasoning, 
applying logical rules over statements can be 
effective
for handling negative queries \cite{asai-hajishirzi-2020-logic,ren2020beta}. 
However, directly manipulating affirmative forms with logic-guided rules may 
fail for commonsense reasoning: 
%
the boundary of commonsense inferences between affirmative and negated statements is not always wholly contrastive. 
Many inferences can be relevant to both forms.
The events ``X puts the potato in the oven'' and ``X doesn't put the potato in the oven,'' could both have an associated inference: ``X wants to make dinner.'' 
The affirmative event clearly implies this inference. 
For the negated event to be worth mentioning on its own \citep{grice1975logic}, an implicit complementary event (\eg, ``X puts the potato in the microwave'') would likely hold, which might validate the inference w.r.t. the negated event. 
To model the defeasibility of commonsense reasoning \citep{Pratt1994,rudinger-etal-2020-thinking}, modeling both common and contrastive inferences of negated forms is necessary. 

\input{tables/data-statistics-table}

%% file: tables/atomic-neg-cue-example-table.tex
\begin{table*}[t]
\small
\renewcommand{\arraystretch}{1.5}
\centering
\begin{tabular}{lll}
\textbf{Types} & \textbf{Example Negation Cues} & \textbf{Example Sentences} \\ \hline

\makecell[tl]{Affixes} & \makecell[tl]{un-, ir-, non-, il-, im-, -less, etc.} & \makecell[tl]{X addresses an \textit{irr}elevant point \\ X is \textit{un}likely to be a spy \\ X \textit{un}saddles the horse} \\

\makecell[tl]{Single-word} & \makecell[tl]{not, no, nothing, nobody, few, little, without, \\never, hardly, rarely, barely, seldomly, etc.} & \makecell[tl]{X does \textit{not} tell the truth to the public \\ X \textit{never} eats ice cream \\ X went to a movie \textit{without} his friends} \\

\makecell[tl]{Multi-word} & \makecell[tl]{no longer, barely/hardly ever, not at all, \\ a lack of, be deprived of, in the absence of, \\ on no condition, by no means, not by any means, \\ under no circumstances, make no attempt to, etc.}
& \makecell[tl]{X \textit{no longer} wants to buy a car \\ X is \textit{not at all} impressed by Y's ideas \\ X \textit{under no circumstances} smokes \\ X is \textit{by no means} cheating on Y} \\ 

\makecell[tl]{Negative Verbs} & \makecell[tl]{oppose, refuse, resist, avoid, disapprove,  \\ lack, discontinue, stop, cease, halt, prohibit, \\ forbid, prevent, reject, fail, etc.} & \makecell[tl]{X \textit{denies} the existence of god \\ X \textit{restrains} himself from eating with Y \\ X \textit{refuses} to be in a relationship} \\ \hline

\end{tabular}
\vspace*{-2mm}
\caption{Negation cues and examples from \dataset{}.}
\vspace*{-2mm}
\label{tab:negation-cue-examples}
\end{table*}

%% file: tables/data-statistics-table.tex
\begin{table*}[t!]
\small
\centering
\begin{tabular}{l|rrrrrr}

\textbf{Type} & & \textbf{$\#$Words} & \textbf{Total}  & \textbf{Train} & \textbf{Development} & \textbf{Test} \\ 
\midrule

\multirow{2}{*}{\atomic} & event & 4.61 & 25,096 & 20,322 & 2,282 & 2,492 \\ 
 & inference & - & 795,059 & 643,571 & 72,227 & 79,261 \\ 
 \midrule

\multirow{2}{*}{\dataset{} - \Logical{} (L)} & event & 4.47 & 8,285 & 4,175 & 1,903 & 2,207 \\ 
 & inference & - & 225,635 & 110,864 & 57,170 & 57,601 \\ 
 \midrule
 
 \multirow{2}{*}{\dataset{} - \Semilogical{} (S)} & event & 4.52 & 5,019 & 2,457 & 1,223 & 1,339 \\ 
 & inference & - & 138,587 & 66,087 & 33,030 & 39,470 \\ 
 \midrule
 
 \multirow{2}{*}{\dataset{} - \Contradiction{} (C)} & event & 4.46 & 9,179 & 3,267 & 2,808 & 3,104 \\ 
 & inference & - & 262,820 & 93,419 & 95,685 & 73,716 \\

\bottomrule
\end{tabular}
\vspace*{-1mm}
\caption{Statistics of \atomic{} and different subsets of \dataset{} (\dataset-L + \dataset-S + \dataset-C).} 
\label{tab:atomic-neg-data-stats}
\end{table*}

%% file: sections/3-resource.tex



 
 



\vspace*{-1mm}
\section{\dataset: Commonsense Inferences of Oppositions and Negations}
\label{sec:resource}

To provide a rich resource of commonsense inferences for opposition and negation events, we design \dataset. Using the same schema as the \atomic{} knowledge graph \citep{sap2019atomic}, we initialize 22,483 negated forms paired to original \atomic{} events and crowdsource 627,042 new inferences for these negated events. Consistent with \atomic{}, \dataset{} is constructed using English formulations of events and inferences. We briefly recap \atomic{} and describe the construction of \dataset{} below.

\paragraph{\atomic{} Background}
The \atomic{} knowledge graph 
contains $\sim$24K base events (\eg, ``X plays the piano'') with 877K accompanying social commonsense inferences (\eg, ``Before, X needs to buy a piano.'') along nine dimensions (\eg, \textit{xNeed}). 
The full description of \atomic{} relation types can be found in Table \ref{tab:atomic-rel-pattern} in the Appendix. 

\subsection{Overview of \dataset{} Construction}

Our knowledge construction pipeline consists of two steps. First, we collect negated and contradictive events by deriving oppositions of events in \atomic{}. Inspired by the distinction made between negation contributed by semantic assertion (explicit negation) or non-asserted content (implicit negation) \cite{XIANG201671}, we define three varieties of negated events: logical negations, semi-logical negations, and commonsense contradictions, which we describe in detail below. Logical and semi-logical negations were heuristically formulated from \atomic{} events. Commonsense contradiction events were crowdsourced from Amazon Mechanical Turk (MTurk). 
Negated events in \dataset{} are assigned to the same data split as the corresponding affirmative event from which they are derived (\eg, negated events for \atomic{} training set events are found in the \dataset{} training set). 

Once a list of negated events is compiled, we crowdsource inferences of these new events on MTurk using similar annotation templates as \citet{sap2019atomic}. We design qualifying tasks to filter out unreliable workers and screen their answers manually for quality control purposes.


\paragraph{Logical Negation} We define logical negation events as events with the negation cue \textit{not} added to their original formulation (\eg, ``X does not play the piano''). However, different positions of the \textit{not} modifier in a clause can result in different \textit{negation scopes}, which can alter the semantics of the event \cite{councill-etal-2010-great}. To be consistent, we systematically insert \textit{not} after the subject of the event clause. 
If necessary, we change verb forms and add auxiliary words (\eg, do, does, did, is, was, can, could, would, should, may, might). 
For quality control, we have human workers validate each logically negated event form  and exclude events that annotators identify as uninterpretable or awkwardly worded. 
For each created event, we then collect the same nine dimensions of inferences as defined in \atomic{}. 
Consequently, we collected 8,285 logically negated events with 225K corresponding inferences (as shown in Table \ref{tab:atomic-neg-data-stats}). Appendix~\ref{sec:appendix:data-collection-details} provides further details of the compilation of logical negation events.

\paragraph{Semi-logical Negation} We define semi-logical negation using explicit cues other than \textit{not}.  We categorize these negation cues (words or phrases) into four subtypes: affixes (\eg, legal/illegal), single-word cues (\eg, never), multi-word cues (\eg, no longer), and negative verbs (\eg, refuse). 
See Table \ref{tab:negation-cue-examples} for examples. 
We create semi-logical negation events by heuristically adding these cues to different positions of \atomic{} events. Similar to logically-negated events, we avoid grammatically incorrect or semantically awkward events by removing auto-generated instances of low quality. The final set of data includes 5,019 semi-logical negation events. We then crowdsource a total of 138K inferences for these new events. 
Appendix~\ref{sec:appendix:data-collection-details} provides further details of the compilation of semi-logical negation events.

\input{tables/atomic-contradiction-example-table}

\paragraph{Commonsense Contradiction} We formulate commonsense contradiction as 
contradictory statements without negation cues. 
Commonsense contradiction events are not identifiable as negations on their own, but demonstrate reversed semantic or pragmatic meaning when paired with their affirmative counterparts (\eg, ``X eats a hamburger'' vs. ``X eats a salad''). To obtain commonsense contradictions, we crowdsource two oppositional events for each \atomic{} event, excluding events with blank placeholders representing generic objects, 
resulting in 40K new commonsense contradiction events. For 9,179 of these events, we crowdsource an additional 262K commonsense inferences. 
Appendix~\ref{sec:appendix:data-collection-details} provides further details of the crowdsourcing of commonsense contradiction events.

%% file: tables/atomic-contradiction-example-table.tex
\begin{table}[t]
\small
\renewcommand{\arraystretch}{1}
\centering
\begin{tabular}{ll}
\textbf{Event} & \textbf{Commonsense Contradiction} \\ 
\toprule


\makecell[tl]{\multirow{2}{*}{X buys a bicycle}} & X buys a car \\
& X donates a bicycle \\
\midrule


\makecell[tl]{\multirow{2}{*}{X walks in the door}} & X stops at the door \\
& X walks out of the building \\
\midrule

\makecell[tl]{\multirow{2}{*}{X works hard all day}} & X plays games all day \\
& X puts in minimal effort all day \\
\midrule

\makecell[tl]{\multirow{2}{*}{X finishes the story}} & X starts the story \\
& X stops halfway through the story \\
\midrule

\makecell[tl]{\multirow{2}{*}{X turns the air blue}} & X secretly curses \\
& X speaks appropriately \\
\bottomrule

\end{tabular}
\vspace*{-1mm}
\caption{Contradictions of events from \atomic{}}
\label{tab:atomic-commonsense-contradiction-pair}
\end{table}

%% file: sections/4-comet-neg.tex
\section{Knowledge Models of Negated Events}
\label{sec:model-comet}

\dataset{} can be used as training data for commonsense models to make inferences about negated events. Here, we recap COMET \cite{bosselut2019comet}, a commonsense knowledge model, and evaluate how training knowledge models on \dataset{} affects their ability to hypothesize commonsense knowledge for negated and oppositional events.

\subsection{Setup}
\label{ssec:model-comet:setup}

Commonsense transformers (COMET) are generative knowledge models that learn to hypothesize commonsense inferences by training on examples from a knowledge graph. Specifically, COMET receives knowledge tuples in $\{h,r,t\}$ form during training, where $h$ is a head entity, $r$ is a relation type, and $t$ is a tail entity. The model is trained to maximize the conditional loglikelihood of predicting the tokens of the tail entity $t$ given the tokens of the head entity $h$ and relation $r$:
\vspace*{-1mm}
\begin{equation}
\mathcal{L}_{G} = - \sum \log{P(t|h, r)}
\label{eq:comet-loss}
\end{equation}


\noindent In \atomic{} and \dataset{}, $h$ corresponds to events, such as ``X has a nightmare,'' $t$ corresponds to commonsense inferences about those events, such as ``X wakes up,'' and $r$ corresponds to commonsense inference types, such as ``As a result, X does...''.





Following  \citet{bosselut2019comet} and \citet{sap2019atomic}, for each event and relation type in \atomic{}, 10 candidate inferences are decoded from COMET using beam search with $b$=10.

\subsection{Experiments}
\label{ssec:model-comet:exps}

As oppositional instances remain challenging to knowledge models such as COMET, we evaluate how \dataset{} can be used to augment the type of examples seen by COMET during training. 

\begin{table}[t]
\small
\renewcommand{\arraystretch}{1.1}
\resizebox{\linewidth}{!}{
\centering
\begin{tabular}{l|lrrrr}

 \textbf{Eval Set} & \textbf{Train Set} & \textbf{PPL $\downarrow$} & \textbf{BL2 $\uparrow$} & \textbf{P@10 $\uparrow$} \\ 
 \midrule

 \multirow{2}{*}{\atomic} & \atomic & 9.30 & \textbf{14.18} & \textbf{55.18} \\
 & \atomic{} + \dataset{} & \textbf{9.28} & 14.05 & *53.61 \\  
 \midrule
 
 \multirow{2}{*}{\makecell[tl]{\dataset-L}} & \atomic & 10.87 & 10.86 & 35.84 \\ 
 & \atomic{} + \dataset{} & \textbf{9.08} & \textbf{11.96} & \textbf{**45.42} \\ 
 \midrule
 
 \multirow{2}{*}{\makecell[tl]{\dataset-S}}& \atomic & 11.69 & 12.07 & 36.89 \\ 
 & \atomic{} + \dataset{} & \textbf{9.80} & \textbf{13.22} & \textbf{**46.88} \\ 
 \midrule
 
 \multirow{2}{*}{\makecell[tl]{\dataset-C}} & \atomic & 12.02 & 14.32 & 46.70 \\ 
 & \atomic{} + \dataset{} & \textbf{11.20} & \textbf{14.64} & \textbf{**50.65} \\ 
 \bottomrule

\hline
\end{tabular}
}
\caption{Evaluations of COMET models trained on \atomic{} and \dataset{} KGs. Training on examples of negated events leads to large improvements in the quality of generated inferences with minimal dropoff in the quality of inferences for affirmative events. 
Single (*) and double asterisks (**) indicate significance at p<0.05 and p<0.01, respectively.} 
\label{tab:comet-neg-result-main}
\end{table}

\paragraph{Evaluation Metrics} Following \citet{bosselut2019comet}, we evaluate the quality of generated inferences using BLEU-2 \cite{bleu-2003} as an automatic evaluation. We also compute the perplexity of models on their reference generations. 

For the human evaluation, we employ human judges from MTurk to identify whether generated commonsense inferences are plausible. 
We randomly sample 100 events from the original \atomic{} test set along with their  
negated counterparts from \dataset. For each event, 
we present every decoded inference to five crowdworkers and ask them to identify whether the inference is plausible given the event. 
For each model trained on a different combination of \atomic{} and \dataset{} (\ie, \dataset-L, \dataset-S, \dataset-C), we evaluate the same events for comparison.
We calculate \Patten{} (\patten{}) across these human ratings, \ie, the average number of correct options per event-relation prompt. Specifically, we average the results from 45K ratings to compute the final human score (100 events $\times$ 9 relations $\times$ 10 options $\times$ 5 annotators). The pairwise agreement score of human evaluation is 63.6, which is on par with other similar commonsense reasoning annotation tasks \cite{rashkin-etal-2016-connotation}.

\paragraph{Does negated event training improve commonsense inference for negated situations?} We train a COMET model on the events from \atomic{} (\ie, COMET-\atomic{}), and another on the examples from both \atomic{} and \dataset{} (\ie, COMET-\full{}). The combined dataset is shuffled so that the original and negated examples are uniformly mixed during training. 

We report our comparison of these two models in Table~\ref{tab:comet-neg-result-main}. The performance of the original COMET model trained only on the \atomic{} knowledge graph 
drops significantly across all types of oppositional instances. Most surprisingly, a drop in performance is also observed on commonsense contradictions (\dataset-C), which have no explicit negation cues. However, commonsense contradiction events can often be richer in content (see Table~\ref{tab:atomic-commonsense-contradiction-pair}), making them more challenging for knowledge models. Meanwhile training on all negated examples in the \dataset{} knowledge graph produces significant improvements across all negation categories (\dataset-\{L,S,C\}), though we do observe a slight drop in human ratings on the examples from the original \atomic{} test set.

\paragraph{Does negated event training deteriorate commonsense inference of affirmative situations?} 
 
We note in Table~\ref{tab:comet-neg-result-main} that training on \atomic{} + \dataset{} hurts inference performance on the original \atomic{} evaluation set. To analyze why COMET-\full{} does not improve on this set of examples, we perform a case study on inferences generated by COMET-\atomic{} and COMET-\full{} under the same event and relation prompt, and note two qualitative patterns. 

First, we observe that COMET-\full{} tends to generate inferences that are less generic, but that may require additional implicit context. 
For example, for the event ``X is really sad'' and the relation \textit{xEffect} (\ie, the effect of the event on X), COMET-\atomic{} generates inferences such as ``cries,'' ``gets depressed'' and ``takes medication.'' 
Conversely, COMET-\full{} generates context-specific inferences such as ``thinks about the past'' and ``thinks about what they did,'' which, while plausible in some context, may be less straightforward when evaluated broadly (not all feelings of sadness lead to reflection on the past or one's own actions). 

Second, we find an overall improvement for certain compositional events in \atomic{} that contain conjunction words: ``and'' or ``but.'' On these examples, COMET-\full{} outperforms COMET-\atomic{} with 12.41 and 12.22 BLEU-2 scores respectively. 
For example, for the event ``X is hot and humid'' and the relation \textit{xEffect}, COMET-\atomic{}’s generation includes correct inferences, such as ``to take a shower,'' ``to cool down,'' ``to drink some water,'' ``to go outside,'' and incorrect inferences, such as ``to turn on the heat'' and ``to drink a hot tea.''
COMET-\full{} generates all of COMET-\atomic{}’s correct inferences, but none of the incorrect inferences, demonstrating that training COMET jointly on \atomic{} and \dataset{} can help avoid incorrect inferences involving commonsense mismatch in more compositional situations. 

In summary, the ability to generate richer, contextual inferences for COMET-\full{} is beneficial when handling complex events, but may not be necessary for many of the simple events in \atomic{}, and may backfire when subtler inferences are made.



\begin{table}[t]
\small
\centering
\begin{tabular}{l|lrrrr}

 \textbf{Eval Set} & \textbf{Train Set} & \textbf{PPL $\downarrow$} & \textbf{BL2 $\uparrow$} & \textbf{P@10 $\uparrow$} \\ 
 \midrule

 \multirow{4}{*}{\atomic} & \atomic & 9.30 & 14.18 & 55.18 \\
 & + \dataset-L & \textbf{9.27} & \textbf{14.20} & \textbf{**58.11} \\ 
 & + \dataset-S & 9.30 & 14.09 &  55.74 \\ 
 & + \dataset-C & 9.29 & 14.10 & **52.22 \\ 
 \midrule
 
 \multirow{4}{*}{\makecell[tl]{\dataset-L}} & \atomic & 10.87 & 10.86 & 35.84 \\ 
 & + \dataset-L & \textbf{9.28} & \textbf{11.94} & \textbf{**44.94} \\ 
 & + \dataset-S & 9.93 & 11.29 & **44.01 \\ 
 & + \dataset-C & 10.34 & 11.04 & **42.33 \\ 
 \midrule
 
 \multirow{4}{*}{\makecell[tl]{\dataset-S}}& \atomic & 11.69 & 12.07 & 36.89 \\ 
 & + \dataset-L & 10.69 & 12.69 & **42.38 \\ 
 & + \dataset-S & \textbf{10.23} & \textbf{12.79} & \textbf{**45.50} \\
 & + \dataset-C & 10.95 & 12.35 & **41.76 \\ 
 \midrule
 
 \multirow{4}{*}{\makecell[tl]{\dataset-C}} & \atomic & 12.02 & 14.32 & 46.70 \\ 
 & + \dataset-L & 11.72 & 14.43 & 47.78 \\ 
 & + \dataset-S & 11.67 & 14.34 & 46.09 \\ 
 & + \dataset-C & \textbf{11.50} & \textbf{14.58} & \textbf{**48.79} \\ 
 \bottomrule

\hline
\end{tabular}
\caption{Ablation results of models trained and evaluated on different portions of \dataset{}. The best result on each subset of \dataset{} comes from training on similar examples. The model trained on negated events from \dataset-L performs the best at generating inferences for the original \atomic{} events. Double asterisks (**) indicate significance at p<0.01.}
\label{tab:comet-neg-result-ablation}
\end{table}

\paragraph{Which variety of negated events are most crucial to include in training sets?} As ablations, we train additional models using different subsets of \dataset{}: \logical~negations (\atomic{} + \dataset-L), \semilogical~negations (\atomic{} + \dataset-S), and commonsense contradictions (\atomic{} + \dataset-C). These ablations evaluate whether knowledge models can adapt to certain types of negation more efficiently with additional data.

In Table \ref{tab:comet-neg-result-ablation}, we show that training with examples of each negation type improves performance on the evaluation set related to that negation type. Interestingly, though, training on certain types of negation examples can also yield benefits downstream on other negation types. For example, training on commonsense contradictions (\dataset-C) provides a clear benefit when evaluating on semi-logically negated events (\dataset-S) as opposed to merely training on \atomic. 
Notably, the knowledge model trained with logically negated examples (\atomic{} + \dataset-L) outperforms the model trained only on \atomic{} on all test sets. 

%% file: sections/5-discriminator.tex
\section{Discriminating Inconsistent Inferences}
\label{sec:model-discriminator}

While training on examples of negated events helps knowledge models generate commonsense inferences for these event types, there is still a large gap compared to their performance on affirmative events. To address this discrepancy, we introduce a discriminator-based approach for distinguishing inconsistent inferences of negated events. Our inference discriminator learns to identify plausible and invalid inferences of events by learning from contrastive samples from \atomic{} and \dataset{}. 

\subsection{Experimental Setup}
\label{ssec:disc:setup}


We fine-tune the RoBERTa-base model \citep{liu2019roberta} as a binary classifier to identify whether a given knowledge tuple $\{h,r,t\}$ is logically valid. The model is trained on 
paired original and negated events as described below. Such training examples inject implicit commonsense nuances that differ between oppositional events to teach the discriminator to identify logical pitfalls. 
Training details for discriminators can be found in Appendix~\ref{sec:appendix:train-discriminator}.

\paragraph{Data}
\label{ssec:disc:data}

The paired events used to train the negation discriminator are automatically constructed from the \atomic{} and \dataset{} knowledge graphs. Positive examples can be constructed by sampling tuples from each knowledge graph. To construct negative training samples, we introduce the concept of \textit{common} and \textit{contrast} sets among inferences of events and their oppositions. 

Common and contrast sets 
distinguish how commonsense inferences are not necessarily negated in the same manner as their corresponding events. While certain inferences of events are also in opposition to a negated event, some may be common. For the events ``X eats a cheeseburger'' and ``X eats a salad,'' an inference such as ``X is hungry'' might be common to both events while inferences such as ``X is unhealthy'' or ``X is healthy'' would be viewed as contrastive. 



Specifically, we assume two head events in \atomic{} and \dataset{}, and their respective set of tail inferences regarding a common relation type. We define the common set of these inferences as the intersection of the two sets of tail inferences connected to each head event by applying the exact match of string forms.
The contrast set is formed by distinct tail inferences connected to the two head events.
Logically valid (\ie, positive) training examples consist of knowledge tuples from \atomic{} and \dataset{}. Logically invalid (\ie, negative) training examples are formed by swapping the set of contrast set inferences between paired original and negated events.\footnote{We note that annotations in \atomic{} and \dataset{} are finite (\ie, not covering the full space of possible commonsense inferences about events). As a result, it is possible that in a more expansive annotation, elements of the contrast sets would in fact be part of the common set of an event and its negation. For the purpose of this work, however, contrast sets were an efficient way of acquiring high-quality semantically negative examples for training discriminators.}

To balance the training set, we sample the same number of positive and negative tuples for original and negation events.
Statistics of the 
resulting training sets are in Table \ref{tab:discriminator-data-stats}.

\begin{table}[t]
\small
\centering
\begin{tabular}{l|c|c}

\textbf{Discriminator} & \textbf{Train Set} & \textbf{Size} \\ 
\midrule

 Logical Negation \textbf{(L)} & \dataset-L & 324,843 \\ 
 Semi-logical Negation \textbf{(S)} & \dataset-S &  194,732 \\ 
 Commonsense Contradiction \textbf{(C)} & \dataset-C &  276,272 \\ 
 All Oppositional Data \textbf{(LSC)} & \dataset & 795,845 \\
\bottomrule
\end{tabular}

\caption{Statistics of data used to train negation discriminators.} 
\label{tab:discriminator-data-stats}
\end{table}

\subsection{Experiments}
\label{ssec:disc_exps}



Using different portions of \dataset{} for training yields four unique discriminators (\ie, \textbf{L}, \textbf{S}, \textbf{C} and \textbf{LSC}) that we apply to commonsense inferences generated by COMET. The discriminators classify each option as either logically \textit{valid} or \textit{invalid}, partitioning the candidates into two sets, which we evaluate with human judgements. As a baseline, we also record the precision of not using a discriminator, 
which assumes all generated inferences are valid candidates (\ie, the \textit{all} set). 
 
\paragraph{Metrics}

We evaluate and compare the quality of the \textit{all}, \textit{valid} and \textit{invalid} sets using BLEU-2 and the same human evaluation as in \S \ref{sec:model-comet}. 
The \textit{all} set contains the full set of 10 candidates, while the \textit{valid} and \textit{invalid} sets have varying number of elements depending on how discriminators classify them, summing to 10. 
To compute statistical significance between valid and all sets, we use a permutation test with 100K permutations. Details 
are provided in Appendix~\ref{app:permutation}.

\begin{table}
\small
\renewcommand{\arraystretch}{1}
\setlength{\tabcolsep}{4.5pt}
\centering
\begin{tabular}{l|lrrrr}

 \textbf{Eval Set} & & \textbf{$\#$} & \textbf{BL2$\uparrow$} & \textbf{P@\textit{k}$\uparrow$} \\ 
 \midrule

 \multirowcell{3}{\atomic}& all & 10.0 & 14.18 & 55.18  \\
 & valid & 6.3 & \textbf{14.24} & \textbf{59.07}  \\ 
 & invalid & 3.7 & 13.93 & 44.10  \\ 
 \midrule
 
 \multirowcell{3}{\dataset-L} & all & 10.0 & 10.86 & 35.84  \\ 
 & valid & 5.6 & \textbf{11.33} & \textbf{45.59}  \\ 
 & invalid & 4.4 & 10.13 & 25.96  \\ 
 \midrule
 
 \multirowcell{3}{\dataset-S} & all & 10.0 & 12.07 & 36.89  \\ 
 & valid & 6.3 & \textbf{12.63} & \textbf{44.93} \\ 
 & invalid & 3.7 & 11.32 & 27.83 \\ 
 \midrule
 
 \multirowcell{3}{\dataset-C} & all & 10.0 & 14.32 & 46.70 \\ 
 & valid & 5.9 & \textbf{14.78} & \textbf{51.45} \\ 
 & invalid & 4.1 & 13.56 & 37.33 \\ 
 \midrule
\end{tabular}
\caption{
The evaluation of the \textit{all}, \textit{valid} and \textit{invalid} sets of inferences generated by COMET-\atomic{} as partitioned by the \textbf{LSC} discriminator. \textbf{P@\textit{k}} corresponds to the human-rated precision of a set. $k$ is the number of elements in \textit{all, valid,} or \textit{invalid} set. For the \textit{valid} set, higher \textbf{P@\textit{k}} is better (\ie, more valid inferences are being partitioned). For the \textit{invalid} set, lower \textbf{P@\textit{k}} is better (\ie, fewer valid inferences are being included). 
}
\label{tab:comet-disc-eval-per-example}
\end{table}

\begin{table}[t!]
\small
\centering
\begin{tabular}{l|l|lc}

\toprule
\textbf{Event + Rel} & \textbf{Generation} & \textbf{V} & \textbf{P}\\
\midrule
 
 \multirow{5}{*}{\makecell[tl]{X does \\not skate \\around\\\textbf{\textit{xAttr}}}} & athletic & \xmark & \xmark \\
 & careless & \xmark & \xmark \\
 & lazy & \cmark & \cmark \\
 & uncoordinated & \cmark & \cmark \\
 & unskilled & \cmark & \cmark \\ 
 \midrule
 
 \multirow{5}{*}{\makecell[tl]{X does \\not sit \\behind Y\\\textbf{\textit{xIntent}}}} & to be alone & \cmark & \cmark \\
 & to be left alone & \cmark & \cmark \\
 & to avoid Y & \cmark & \cmark \\
 & to sit & \xmark & \xmark \\
 & to wait & \cmark & \xmark \\
 \midrule

 \multirow{5}{*}{\makecell[tl]{X does \\not look \\angry\\\textbf{\textit{xNeed}}}} & to calm down & \xmark & \cmark \\
 & to watch a movie & \cmark & \xmark \\
 & to have been provoked & \xmark & \xmark \\
 & to not be angry & \cmark & \cmark \\
 & to be calm & \cmark & \cmark \\
 \midrule

 \multirow{5}{*}{\makecell[tl]{X refuses \\to hear a \\scary noise\\\textbf{\textit{xWant}}}} & to run away & \xmark & \xmark \\
 & to go to sleep & \cmark & \cmark \\
 & to be safe & \cmark & \cmark \\
 & to keep quiet & \cmark & \cmark \\
 & to avoid the noise & \cmark & \cmark \\
 \midrule
 
 \multirow{5}{*}{\makecell[tl]{X never \\brings Y into \\conflicts\\\textbf{\textit{oWant}}}} & to avoid X & \xmark & \xmark \\
 & to be left alone & \xmark & \cmark \\
 & to thank X & \cmark & \cmark \\
 & to fight back & \xmark & \xmark \\
 & to avoid conflict & \xmark & \cmark \\
 \midrule
 
 \multirow{5}{*}{\makecell[tl]{X scarcely\\gets sunburned\\ \\\textbf{\textit{xReact}}}} & burned & \xmark & \xmark\\
 & hurt & \xmark & \xmark\\
 & sick & \xmark & \xmark\\
 & sad & \xmark & \xmark\\
 & satisfied & \cmark & \cmark\\
 \midrule
 
 \multirow{5}{*}{\makecell[tl]{X under no\\ circumstances\\forgets Y's wallet\\\textbf{\textit{oReact}}}} & upset & \xmark & \xmark \\
 & sad & \xmark & \xmark \\
 & angry & \xmark & \xmark \\
 & thankful & \cmark & \cmark \\
 & grateful & \cmark & \cmark \\
 \midrule
 
 \multirow{5}{*}{\makecell[tl]{X has trouble \\with advertising \\X's business\\\textbf{\textit{xEffect}}}} & loses money & \cmark & \cmark \\
 & loses clients & \cmark & \cmark \\
 & gets fired & \cmark & \cmark \\
 & gets sued & \xmark & \xmark \\
 & cries & \cmark & \cmark \\
 \midrule
 
 \multirow{5}{*}{\makecell[tl]{X puts Y \\out of mind \\ \\\textbf{\textit{oEffect}}}} & has a better day & \xmark & \xmark \\
 & becomes sad & \cmark & \cmark \\
 & cries & \cmark & \cmark \\
 & is grateful towards X & \xmark & \xmark \\
 & feels better & \xmark & \xmark \\
\bottomrule

\end{tabular}
\vspace*{-1mm}
\caption{
Inferences of randomly selected \dataset{} events by COMET-\atomic{}. 
The top 5 options are classified as \textit{valid} or \textit{invalid} by the \textbf{LSC} discriminator. \textbf{V} indicates whether an option is classified as \textit{valid} by the \textbf{LSC} discriminator. \textbf{P} indicates whether an option is plausible judging by humans.} 
\vspace*{-2mm}
\label{tab:discriminator-generations-all}
\end{table}


\paragraph{Do discriminators effectively distinguish inconsistent inferences?}
The results in Table \ref{tab:comet-disc-eval-per-example} demonstrate that the discriminator trained on all subsets of \dataset{} (\textbf{LSC}) can select subsets of inferences (\ie, the \textit{valid} set) that are more logically consistent with their seed event. This observation holds across all evaluation subsets of \dataset{}, as well as the original \atomic{} evaluation set. 
Table \ref{tab:discriminator-generations-all} shows examples of \textit{valid} and \textit{invalid} candidates for negated and contradicted events from \dataset{} as specified by the \textbf{LSC} discriminator. The discriminator is notably good at identifying invalid inferences wrongly associated to corresponding affirmative events (\eg, ``athletic'' and ``careless'' for the event ``X does not skate around'' under the relation, \textit{xAttr}).

However, this analysis leaves open the possibility that we are generating too many inferences for each event, but that the decoder could rank correct inferences higher among the full set of generated candidates. To evaluate this possibility, we count the number of elements in the \textit{valid} sets for each example and only keep the same number of the top-scoring elements from the \textit{all} set (scored using generation perplexity). In Table~\ref{tab:o-p@len(valid)}, we see the average precision score for the pruned \textit{all} sets (\patlenvalid) 
still underperforms the precision of their corresponding \textit{valid} sets.
\begin{table}[t]
\footnotesize
\renewcommand{\arraystretch}{1}
\setlength{\tabcolsep}{4.5pt}
\centering
\begin{tabular}{l|r|rrrr}
 
 \multicolumn{2}{c|}{\diagbox[width=\widthof{xxxxxxxxxx}]{\textbf{Eval}}{\textbf{Disc}}} & \textbf{L} & \textbf{S} & \textbf{C} & \textbf{LSC} \\ 
 \midrule

 \multirowcell{3}{\atomic} & all & \textbf{55.69} & 55.93 & 56.94 & 58.30 \\ 
 & valid & 55.65 & \textbf{56.18} & \textbf{57.26} & \textbf{59.07} \\ 
 & \%iprv & -0.07 & 0.44 & 0.57 & \underline{1.32} \\ 
 \midrule
 
 \multirowcell{3}{\dataset-L} & all & 39.46 & 37.85 & 36.43 & 39.45 \\ 
 & valid & \textbf{**46.39} & \textbf{**41.93} & \textbf{37.54} & \textbf{**45.59} \\ 
 & \%iprv & \underline{17.55} & 10.78 & 3.03 & 15.57 \\ 
 \midrule
 
 \multirowcell{3}{\dataset-S} & all & 37.13 & 39.29 & 37.72 & 38.55 \\ 
 & valid & \textbf{37.48} & \textbf{**44.58} & \textbf{39.03} & \textbf{**44.93} \\ 
 & \%iprv & 0.96 & 13.47 & 3.45 & \underline{16.56} \\ 
 \midrule
 
 \multirowcell{3}{\dataset-C} & all & \textbf{46.92} & 47.32 & 48.26 & 48.81 \\ 
 & valid & 46.83 & \textbf{47.68} & \textbf{48.79} & \textbf{*51.45} \\ 
 & \%iprv & -0.20 & 0.75 & 1.09 & \underline{5.40} \\ 
 \midrule
\end{tabular}
\vspace*{-2mm}
\caption{\patlenvalid{} scores of the \textit{all} and \textit{valid} sets determined by the \textbf{L}, \textbf{S}, \textbf{C} and \textbf{LSC} discriminators. 
Generations are from COMET-\atomic.
Asterisks (**) indicate significance at p<0.01. iprv\% is the improvement of the \textit{valid} over the \textit{all} set. \underline{Underlines} show the highest iprv\% across discriminators.}
\label{tab:o-p@len(valid)}
\end{table}

\paragraph{Which negation categories are most important to provide a discriminator for?} To examine the generalization effects of each negation type, we also train discriminators on a single negation subset of \dataset{} examples (\ie, \textbf{L}, \textbf{S}, \textbf{C}) and compare the \patlenvalid{} score of the \textit{all} and \textit{valid} sets. 
Results in Table \ref{tab:o-p@len(valid)} indicate that each discriminator is best for identifying valid inferences for the types of events on which it was trained. The \textbf{L}, \textbf{S}, and \textbf{C} discriminators all achieve improvements when partitioning events similar to their training. 
However, the \textbf{LSC} discriminator trained on all negation forms shows the largest \textit{valid} set improvement across all discriminators on \atomic{}, \dataset-S, and \dataset-C. On \dataset-L, the \textbf{LSC} discriminator still yields a significantly improved \textit{valid} set.

%% file: sections/6-discussion.tex
\section{Discussion}
\label{sec:discussion}

\begin{table}[t]
\footnotesize
\renewcommand{\arraystretch}{1}
\setlength{\tabcolsep}{4.5pt}
\centering
\begin{tabular}{l|r|rrrr}
 
 \multicolumn{2}{c|}{\diagbox[width=\widthof{xxxxxxxxxx}]{\textbf{Eval}}{\textbf{Disc}}} & \textbf{L} & \textbf{S} & \textbf{C} & \textbf{LSC} \\ 
 \midrule

 \multirowcell{3}{\atomic} & all & 54.16 & 54.49 & 55.03 & 55.68 \\ 
 & valid & \textbf{54.20} & \textbf{54.64} & \textbf{55.71} & \textbf{**57.58} \\ 
 & \%iprv & 0.08 & 0.28 & 1.23 & \underline{3.41} \\ 
 \midrule
 
 \multirowcell{3}{\dataset-L} & all & 46.54 & 46.26 & 46.15 & 46.39 \\ 
 & valid & \textbf{**50.71} & \textbf{**48.36} & \textbf{46.16} & \textbf{**49.85} \\ 
 & \%iprv & \underline{8.98} & 4.54 & 0.03 & 7.45 \\ 
 \midrule
 
 \multirowcell{3}{\dataset-S} & all & 46.90 & 47.73 & 47.47 & 47.53 \\ 
 & valid & \textbf{47.14} & \textbf{**50.42} & \textbf{48.20} & \textbf{**50.62} \\ 
 & \%iprv & 0.51 & 5.65 & 1.55 & \underline{6.50} \\ 
 \midrule
 
 \multirowcell{3}{\dataset-C} & all & 50.80 & 51.29 & 51.28 & 51.83 \\ 
 & valid & \textbf{50.94} & \textbf{51.52} & \textbf{52.65} & \textbf{*53.91} \\ 
 & \%iprv & 0.28 & 0.45 & 2.67 & \underline{4.02} \\ 
 \midrule
\end{tabular}
\vspace*{-2mm}
\caption{\patlenvalid{} scores of the \textit{all} and \textit{valid} sets determined by the \textbf{L}, \textbf{S}, \textbf{C} and \textbf{LSC} discriminators. Generations are from COMET-\full{}
. Single (*) and double asterisks (**) indicate significance at p<0.05 and p<0.01, respectively. iprv\% is the improvement of the \textit{valid} over the \textit{all} set. \underline{Underlines} indicate the highest iprv\% across discriminators.}
\label{tab:o+l+s+p-p@len(valid)}
\end{table}

\paragraph{Are learning-based and discriminator-based approaches 
complementary?}

We apply our discriminators to the generations of the COMET model trained on \dataset{}. 
In Table \ref{tab:o+l+s+p-p@len(valid)}, we see that the \textbf{LSC} discriminator, when applied to generations of COMET  trained on \dataset{}, achieves significant improvements over all evaluation sets, including the original events. 
The full evaluation of the \patlenvalid{} and \patthree{} scores of applying different discriminators to generations of COMET trained on different data over all evaluation sets are shown in Table \ref{tab:full-p@len(valid)} and \ref{tab:full-p@3} in Appendix \ref{sec:appendix}.

\begin{table}
\small
\renewcommand{\arraystretch}{0.8}
\centering
\begin{tabular}{l|l|rrr}

 \textbf{Beam Size} & \textbf{Set} & \textbf{$\#$\cmark} & \textbf{$\#$total} & \textbf{P@$\#$total} \\
 \toprule

  \multirow{2}{*}{\textbf{10}} & all & 3.6 & 10.0 & 35.84 \\
 & valid & 2.1 & 4.4 & 45.59 \\ 
 \cmidrule{0-1}
 
 \multirow{2}{*}{\textbf{25}} & all & 8.1 & 25.0 & 32.29 \\
 & valid & \textbf{4.3} & 10.5 & \textbf{38.18} \\ 
 \bottomrule

\end{tabular}

\caption{Number of correct generations from applying the \textbf{LSC} discriminator to generations of COMET-\atomic{}
for beam size of 10 and 25 for logical negation events. \textit{$\#$\cmark} is the number of correct options. \textit{$\#$total} is the number of options in each set.}
\label{tab:valid-option-yield}
\end{table}



 



\vspace*{2mm} 
\noindent \textbf{Can discriminators be used to more aggressively generate inferences?} While applying discriminators to generated inferences yields a \textit{valid} subset with higher accuracy, we are left with fewer correct inferences in total. Thus, we investigate the efficiency of using discriminators to expand the number of inferences generated. We decode inferences from COMET  
with beam size 25, and then apply the discriminator to this larger candidate set. 

Table \ref{tab:valid-option-yield} shows that for logical negation, the \textit{valid} set of beam 25 has higher accuracy and more correct options than the \textit{all} set of beam 10. Thus, when we have a larger and potentially more noisy set of candidates, applying the negation discriminator yields a set of options that have higher quality than using all the candidates from a smaller set of initial generations.

%% file: sections/8-conclusion.tex
\section{Conclusion}
\label{sec:conclusion}

We present the first comprehensive study on commonsense implications of negations and contradictions. To expand commonsense resources for the challenge of negation modeling, we introduce \dataset, a large scale commonsense knowledge graph for negated and contradicted events. We use \dataset{} to train commonsense knowledge models and demonstrate that it effectively enriches machine commonsense inference capabilities around negation.
Lastly, we propose a negation discriminator  
capable of identifying logical flaws in commonsense inferences. 
By combining the model trained on
\dataset{} with the negation discriminator, we achieve a further performance boost.

%% file: sections/ethical.tex
\section*{Ethical Considerations}

\subsection*{\dataset{} Language Choice and Implications}

We select English as the base language of \dataset{} so that our resource may be directly linked with the original \atomic{} knowledge graph. We acknowledge, however, that resources in English are more likely to reflect the mindsets and behaviors of English speakers.  Furthermore, and in our case specifically, our annotators were primarily from the US. Consequently, this language choice biases the content of the knowledge graph toward North American perspectives, which affects what models trained on these resources would learn about social norms \citep{acharya2020atlas}. Future works may also include other languages and cultures to make the \dataset{} resource more culturally and ideologically inclusive.

\subsection*{Crowdworker Recruitment, Quality Control and Remuneration}

We recruit crowdworkers from MTurk who are located within the US with HIT approval rates higher than 98\%. 
To ensure high quality task completions, we post pilot batches and manually examine tens of thousands of responses to identify users who provide high quality annotations. 
We select 834 qualified users for the formal data collection and human evaluation tasks. 
Since the entire study spans multiple months, we regularly sample responses to re-examine their quality during the formal study, and remove HITs from crowdworkers who provide decreased-quality responses over time. 
We are particularly cautious about the human evaluation tasks, so even with qualified users, we still comprehensively examine tens of thousands of human evaluation tasks by grouping HITs per users, and look at their responses together to identify potential spamming behaviors and inconsistencies.

For the data collection and human evaluation tasks, we aimed to compensate crowdworkers with an average of \$15 per hour. To ensure a fair payment, we first post a pilot task to evaluate average time cost of a specific task, and pay users at a high rate in this round to avoid underpayment during the pilot study. We then calculate new payment from the pilot task such that approximately 75\% of the HITs would have been paid with more than \$15 per hour at the adjusted rate in the pilot round. We then adopt this new rate for the formal study. We repeat the above procedure of determining payment periodically during the study to ensure the crowdworkers are consistently well-paid.

%% file: sections/appendix.tex
\appendix
\newpage
\newpage

\section{Appendices}
\label{sec:appendix}

\subsection{\dataset{} Data Collection Details}
\label{sec:appendix:data-collection-details}

\paragraph{Heuristic of Creating Logical and Semi-logical Negation Events}

For logical negation, with the majority of the original events being simple sentences with one predicate, our general rule of thumb is to negate the original event at the sentence level. Specifically, with respect to each original event, we first identify each tokens’ part of speech (POS) tags via the NLTK toolkit\footnote{\url{https://www.nltk.org}}. Then, we insert the negation cue \textit{not} after the subject of each sentence, with majority of the case the entity ``PersonX,'' with few exceptions of ``PersonX’s'' and ``PersonX and PersonY.''

To ensure the grammar correctness of the heuristically generated logical negation events, we add appropriate auxiliary verbs (\eg, do, does, did, is, was, can, could, would, should, may, might) in accordance with the tenses (\eg, present, past, future) of the original events. 
Since NLTK’s POS parser fails to recognize some of the verbs that have both noun and verb usage (\eg, ``waters'' the plant, ``supports'' her argument), we curate a list of dual-used words and map them manually. 
Also, while converting the original events to their logical negation counterparts, we revise grammar mistakes from \atomic{} and exclude awkward expressions as much as possible. 
In addition, to make the negation forms sound more natural, we replace the modifier ``some'' by ``any'' during conversion (\eg, ``PersonX buys some shoes'' is converted to ``PersonX doesn't buy any shoes''). 
For the minority of compound events with clauses or complex sentence structures, we disregard them for the purpose of ensuring the data quality.

For semi-logical negation events, we curate a list of semi-logical negation cues besides \textit{not} from various sources\footnote{\url{https://dictionary.cambridge.org/us/grammar/british-grammar/negation_2}} \cite{councill-etal-2010-great, hossain2020its, kim-etal-2019-probing} and categorize them into four types including affixes, single-word cues, multi-word cues and negative verbs (Table \ref{tab:negation-cue-examples}). We identify appropriate rules to insert each semi-logical negation cue in simple base events from \atomic{} consisting of a subject and a predicate. We apply the rules to original events from \atomic{} and randomly select at least 200 automatically generated semi-logical negation events per each negation cue for manual screening by the first author to avoid misplacement of negation cues and awkward expressions. In the end, we were able to identify 5,019 high quality semi-logical negation events originating from \atomic{}.

As a final quality control step of the constructed logical and semi-logical events, after obtaining the crowdsourced inferences for each event, we remove all events that annotators comment as ``unclear,'' ``doesn’t make sense'' or ``grammatically wrong.''

\paragraph{Crowdsourcing of Commonsense Contradiction Events}

\begin{figure*}[t]
    \centering
    \includegraphics[width=1\textwidth]{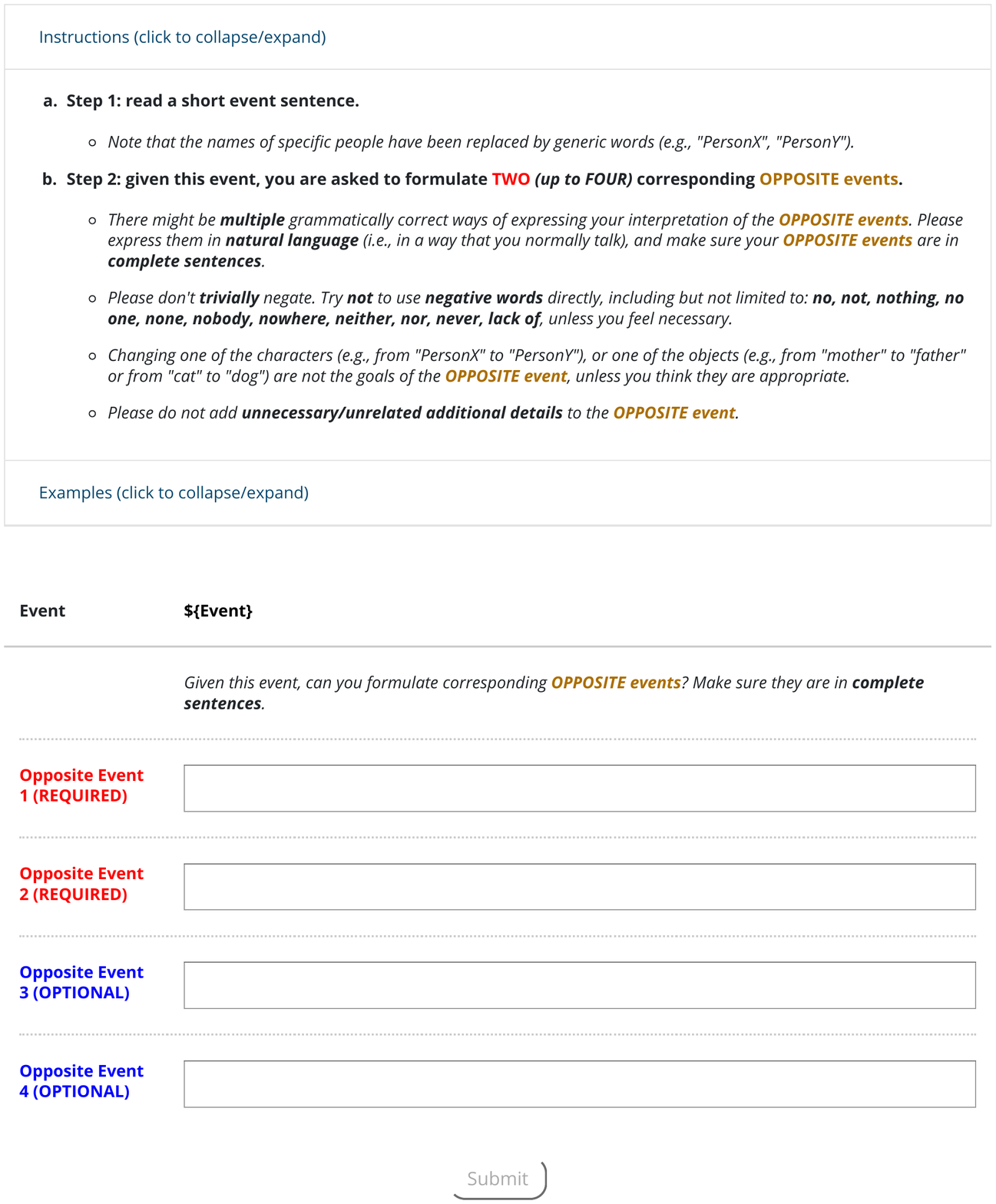}
    \caption{Snippet of the annotation task used to collect commonsense contradiction events.}
    \label{fig:commonsense_contradiction_amt_hit}
\end{figure*}

For collecting commonsense contradiction events, we present an original \atomic{} event to the annotators and ask them to formulate corresponding opposite events. We exclude \atomic{} events with placeholders representing generic objects) to capture semantic and pragmatic subtlety. In the MTurk task, we present annotators detailed instructions of formulating the opposite events (\eg, avoid using negative words as much as possible, use complete sentences, follow grammar rules) and concrete examples as references. 
Figure \ref{fig:commonsense_contradiction_amt_hit} shows details of the MTurk task.
Although we explicitly instruct annotators to avoid using negation cues, there are still some exceptions. 
Therefore, after the compilation of all commonsense contradiction events, we remove ones that contain any explicit negation cues to make sure the categorization is clean.

\paragraph{Crowdsourcing of \dataset{} Event Inferences}

For the collection of \dataset{} event inferences, we adopt the MTurk templates used by the original \atomic{} data collection\footnote{\url{https://homes.cs.washington.edu/~msap/atomic/mTurkFiles/}}.
Similarly to logical and semi-logical events, we remove all inferences of events that annotators comment as ``unclear,'' ``doesn’t make sense'' or ``grammatically wrong.''

\subsection{Training Details of COMET Models}

\paragraph{Input}

A knowledge tuple $\{h,r,t\}$ is represented as a concatenated sequence with tokens of each element in the tuple: $X = \{X^{h}, X^{r}, X^{t}\}$ where $X^{h} = \{x_{0}^{h},...,x_{|h|}^{h}\}$ are the tokens comprising the event, $X^{r} = \{x_{0}^{r},...,x_{|r|}^{r}\}$ as tokens comprising the relation, and $X^{t} = \{x_{0}^{t},...,x_{|t|}^{t}\}$ are the tokens comprising the commonsense inference. 

\paragraph{Initialization}
Similar to \citet{bosselut2020dynamic}, we initialize the trained parameters of COMET to the 345M parameter GPT2 model (GPT2-M) from \citet{Radford2019LanguageMA}. 
Special tokens that represent relation types (\eg, \textit{xIntent}) are added to the vocabulary and initialized via sampling from the normal distribution.

\paragraph{Hyperparameters}

Following \citet{bosselut2019comet}, we use a dropout rate of 0.1 and GeLU \cite{hendrycks2020gaussian} units as activation functions. During training, we use the Adam optimizer \cite{kingma2017adam} with a batch size of 64. For COMET models trained on different subsets of the \atomic{} and \dataset{} datasets, we adopt a maximum learning rate of 6.25e-5 with a warmup period of 0.002 times of the total number of minibatches customized for each model, which decays linearly until finishing training. 

We train different COMET models for different subsets of the full  on original data (\atomic), original and logical negation data (\atomic{} + \dataset-L), original and semi-logical negation data (\atomic{} + \dataset-S), original and commonsense contradiction data (\atomic{} + \dataset-C), and the overall dataset (\atomic{} + \dataset), for 21k, 25K, 24K, 24K and 29K minibatches respectively, and apply early stopping for all models. The rest of the hyperparameters are the same as those of GPT2-M in \citet{Radford2019LanguageMA} implemented via the publicly available HuggingFace API\footnote{https://huggingface.co/transformers/} .

All models are fine-tuned and evaluated on a single NVIDIA QUADRO RTX 8000 GPU for six to twelve hours depending on the complexity of the experimental setup.

\subsection{Training Details of Negation Discriminator}
\label{sec:appendix:train-discriminator}

\paragraph{Input}

As input to the discriminator model, we design sentence patterns that express relation types in natural language and fill out the patterned sentences with events and conditions before encoding them (\eg, ``PersonX addresses a talk. As a result, PersonX wants to convince others.''). Relations and their corresponding patterned sentences are listed in Table \ref{tab:atomic-rel-pattern}. Adopting patterned sentences is found to be a more effective approach than concatenating components in knowledge tuples from the pilot study.

\paragraph{Loss Function}

The negation discriminator is trained to minimized the binary cross-entropy loss:
\vspace*{-2mm}
\begin{equation}
\mathcal{L}_{D} = y \cdot \log{P( y)} + (1-y) \cdot \log{(1-P( y))}
\label{eq:disc-loss}
\end{equation}

\noindent where $y$ is the label for an input (\ie, logically valid or invalid).
 
\paragraph{Hyperparameters}

Parameters are initialized with the trained weights of the RoBERTa-base model in \citet{liu2019roberta}. 
During training, we use the Adam optimizer \cite{kingma2017adam} and train the model with a batch size of 64. We adopt a maximum learning rate of 4.5e-5 with a warmup period of 10 minibatches. We trained \textbf{L}, \textbf{S}, \textbf{C}, \textbf{LSC} discriminators, for 25K, 14K, 21K and 6K minibatches respectively, and apply early stopping for all models. We use a probability threshold of 0.7 to determine whether an input knowledge tuples to the discriminator is plausible based on pilot study on the development sets. The rest of the hyperparameters are the same as those of RoBERTa-base \cite{liu2019roberta} implemented via the publicly available HuggingFace API\footnote{https://huggingface.co/transformers/} .

All models are fine-tuned and evaluated on a single NVIDIA QUADRO RTX 8000 GPU for four to six hours depending on the different experimental setups.

\begin{table}[t]
\small
\centering
\begin{tabular}{l|l}
 \textbf{Relation} & \textbf{Patterned sentences} \\ 
 \midrule
 \textit{xIntent} & $\{h\}$. Because PersonX wanted $\{t\}$.\\
 \textit{xNeed} & $\{h\}$. Before, PersonX needed $\{t\}$. \\
 \textit{xAttr} & $\{h\}$. PersonX is seen as $\{t\}$. \\ 
 \textit{xWant} & $\{h\}$. As a result, PersonX wants $\{t\}$. \\ 
 \textit{oWant} & $\{h\}$. As a result, others want $\{t\}$. \\ 
 \textit{xEffect} & $\{h\}$. As a result, PersonX then $\{t\}$. \\ 
 \textit{oEffect} & $\{h\}$. As a result, others then $\{t\}$. \\ 
 \textit{xReact} & $\{h\}$. As a result, PersonX feels $\{t\}$. \\ 
 \textit{oReact} & $\{h\}$. As a result, others feel $\{t\}$. \\
 \bottomrule
\end{tabular}

\caption{Patterned sentences representing relation types in \atomic{}, used to construct inputs for training negation discriminators.} 
\label{tab:atomic-rel-pattern}
\end{table}

\begin{table*}
\footnotesize
\renewcommand{\arraystretch}{1}
\setlength{\tabcolsep}{4.5pt}
\centering
\begin{tabular}{l|r|rrr|rrr|rrr|rrr}
 
 \multicolumn{2}{l|}{\textbf{Eval}} & \multicolumn{3}{c|}{\textbf{\atomic}} & \multicolumn{3}{c|}{\textbf{\dataset-L}} & \multicolumn{3}{c|}{\textbf{\dataset-S}} &  \multicolumn{3}{c}{\textbf{\dataset-C}} \\
 \toprule
  
 \textbf{Trn} & \textbf{Dis} & all & valid & ipv\% &  all & valid & ipv\% &  all & valid & ipv\% & all & valid & ipv\%  \\ 
 \midrule

 \multirow{4}{*}{\atomic} & L & \textbf{55.69} & 55.65 & -0.07 & 39.46 & \textbf{**46.39} & 17.55 & 37.13 & \textbf{37.48} & 0.96 & \textbf{46.92} & 46.83 & -0.20 \\ 
 & S & 55.93 & \textbf{56.18} & 0.44 & 37.85 & \textbf{**41.93} & 10.78 & 39.29 & \textbf{**44.58} & 13.47 & 47.32 & \textbf{47.68} & 0.75 \\ 
 & C & 56.94 & \textbf{57.26} & 0.57 & 36.43 & \textbf{37.54} & 3.03 & 37.72 & \textbf{39.03} & 3.45 & 48.26 & \textbf{48.79} & 1.09 \\ 
 & LSC & 58.30 & \textbf{59.07} & 1.32 & 39.45 & \textbf{**45.59} & 15.57 & 38.55 & \textbf{**44.93} & 16.56 & 48.81 & \textbf{*51.44} & 5.40 \\  
 \cmidrule{1-2}

 \multirow{4}{*}{\makecell[tl]{\atomic+\\\dataset-L}} & L & 58.62 & \textbf{58.72} & 0.16 & 46.05 & \textbf{**51.19} & 11.16 & 42.42 & \textbf{42.89} & 1.11 & \textbf{47.98} & 47.97 & -0.02 \\ 
 & S & 58.93 & \textbf{59.31} & 0.64 & 45.90 & \textbf{**49.00} & 6.77 & 44.22 & \textbf{**47.59} & 7.64 & 48.10 & \textbf{48.69} & 1.24 \\ 
 & C & 59.63 & \textbf{60.07} & 0.73 & 45.88 & \textbf{46.23} & 0.76 & 43.40 & \textbf{43.74} & 0.79 & 48.81 & \textbf{49.84} & 2.12 \\ 
 & LSC & 60.83 & \textbf{62.49} & 2.74 & 45.96 & \textbf{**50.19} & 9.20 & 44.61 & \textbf{**48.30} & 8.27 & 49.73 & \textbf{*51.97} & 4.51 \\  
 \cmidrule{1-2}
 
 \multirow{4}{*}{\makecell[tl]{\atomic+\\\dataset-S}} & L & \textbf{56.37} & 56.35 & -0.05 & 44.77 & \textbf{**51.76} & 15.60 & 45.58 & \textbf{45.87} & 0.63 & 46.24 & \textbf{46.29} & 0.11 \\ 
 & S & 56.60 & \textbf{56.66} & 0.11 & 44.39 & \textbf{**47.42} & 6.83 & 46.07 & \textbf{**48.32} & 4.89 & 46.62 & \textbf{47.17} & 1.19 \\ 
 & C & 57.46 & \textbf{57.60} & 0.23 & 44.46 & \textbf{45.39} & 2.07 & 45.81 & \textbf{47.15} & 2.93 & 47.38 & \textbf{48.83} & 3.06 \\ 
 & LSC & 58.74 & \textbf{*60.39} & 2.81 & 44.94 & \textbf{**49.88} & 10.98 & 46.08 & \textbf{**48.67} & 5.62 & 48.56 & \textbf{**51.22} & 5.46 \\  
 \cmidrule{1-2}
 
 \multirow{4}{*}{\makecell[tl]{\atomic+\\\dataset-C}} & L & 52.72 & \textbf{52.73} & 0.02 & 43.45 & \textbf{**49.62} & 14.20 & 41.83 & \textbf{41.88} & 0.12 & 48.93 & \textbf{48.97} & 0.07 \\ 
 & S & 52.93 & \textbf{53.33} & 0.76 & 42.66 & \textbf{**46.40} & 8.75 & 42.57 & \textbf{**46.40} & 8.98 & 49.18 & \textbf{49.49} & 0.62 \\ 
 & C & 53.70 & \textbf{54.07} & 0.69 & 42.83 & \textbf{43.26} & 1.00 & 42.25 & \textbf{42.70} & 1.07 & 49.30 & \textbf{*50.97} & 3.38 \\ 
 & LSC & 54.38 & \textbf{55.74} & 2.49 & 44.17 & \textbf{**48.84} & 10.58 & 42.37 & \textbf{**46.22} & 9.10 & 50.07 & \textbf{**52.80} & 5.46 \\   
 \cmidrule{1-2}
 
 \multirow{4}{*}{\makecell[tl]{\atomic+\\\dataset}} & L & 54.16 & \textbf{54.20} & 0.08 & 46.54 & \textbf{**50.71} & 8.98 & 46.90 & \textbf{47.14} & 0.51 & 50.80 & \textbf{50.94} & 0.28 \\ 
 & S & 54.49 & \textbf{54.64} & 0.28 & 46.26 & \textbf{**48.36} & 4.54 & 47.73 & \textbf{**50.42} & 5.65 & 51.29 & \textbf{51.52} & 0.45 \\ 
 & C & 55.03 & \textbf{55.71} & 1.23 & 46.15 & \textbf{46.16} & 0.03 & 47.47 & \textbf{48.20} & 1.55 & 51.28 & \textbf{52.65} & 2.67 \\ 
 & LSC & 55.68 & \textbf{**57.58} & 3.41 & 46.39 & \textbf{**49.85} & 7.45 & 47.53 & \textbf{**50.62} & 6.50 & 51.83 & \textbf{*53.91} & 4.02 \\  
 \bottomrule
\end{tabular}

\caption{For generations of COMET models trained on different subsets of \atomic{} and \dataset{}, the \Patlenvalid{} scores of the \textit{all} and \textit{valid} sets determined by \textbf{L}, \textbf{S}, \textbf{C} and \textbf{LSC} discriminators with respect to the original and negation evaluation sets. The single (*) and double asterisks (**) indicate significance at p<0.05 and p<0.01 respectively. iprv\% is the percentage improvement of the \textit{valid} set over the \textit{all} set.}
\label{tab:full-p@len(valid)}
\end{table*}

\begin{table*}
\footnotesize
\renewcommand{\arraystretch}{1}
\setlength{\tabcolsep}{4.5pt}
\centering
\begin{tabular}{r|r|rrr|rrr|rrr|rrr}
 
 \multicolumn{2}{l|}{\textbf{Eval}} & \multicolumn{3}{c|}{\textbf{\atomic}} & \multicolumn{3}{c|}{\textbf{\dataset-L}} & \multicolumn{3}{c|}{\textbf{\dataset-S}} &  \multicolumn{3}{c}{\textbf{\dataset-C}} \\
 \toprule
  
 \textbf{Trn} & \textbf{Dis} & all & valid & iprv\% & all & valid & iprv\% &  all & valid & iprv\% &  all & valid & iprv\%  \\ 
 \midrule

 \multirow{4}{*}{\makecell[tl]{\atomic}} & L & 59.41 & \textbf{59.65} & 0.40 & 44.92 & \textbf{**49.95} & 11.20 & 39.47 & \textbf{39.94} & 1.21 & 50.77 & \textbf{50.91} & 0.27 \\ 
 & S & 59.48 & \textbf{60.14} & 1.12 & 42.88 & \textbf{**46.24} & 7.83 & 45.27 & \textbf{**49.25} & 8.81 & 51.22 & \textbf{51.84} & 1.21 \\ 
 & C & 59.89 & \textbf{60.89} & 1.66 & 39.20 & \textbf{40.28} & 2.75 & 40.07 & \textbf{41.40} & 3.32 & 51.77 & \textbf{52.88} & 2.15 \\ 
 & LSC & 61.37 & \textbf{63.12} & 2.85 & 46.00 & \textbf{**50.34} & 9.44 & 46.15 & \textbf{**50.23} & 8.85 & 53.29 & \textbf{55.24} & 3.65 \\ 
 \cmidrule{1-2}
 
 \multirow{4}{*}{\makecell[tl]{\atomic+\\\dataset-L}} & L & 61.33 & \textbf{61.57} & 0.39 & 51.47 & \textbf{**56.16} & 9.11 & 45.73 & \textbf{46.04} & 0.68 & 51.40 & \textbf{51.57} & 0.33 \\ 
 & S & 61.13 & \textbf{62.05} & 1.51 & 50.12 & \textbf{**53.40} & 6.54 & 50.84 & \textbf{**54.09} & 6.40 & 51.89 & \textbf{52.91} & 1.96 \\ 
 & C & 61.48 & \textbf{62.96} & 2.40 & 48.23 & \textbf{49.06} & 1.72 & 46.42 & \textbf{46.99} & 1.22 & 52.31 & \textbf{53.67} & 2.61 \\ 
 & LSC & 63.66 & \textbf{*65.85} & 3.44 & 51.90 & \textbf{**56.26} & 8.40 & 51.12 & \textbf{**54.59} & 6.78 & 53.97 & \textbf{56.15} & 4.04 \\ 
 \cmidrule{1-2}
 
 \multirow{4}{*}{\makecell[tl]{\atomic+\\\dataset-S}} & L & 60.25 & \textbf{60.65} & 0.67 & 48.11 & \textbf{**54.45} & 13.18 & 45.97 & \textbf{46.35} & 0.81 & 50.82 & \textbf{50.89} & 0.15 \\ 
 & S & 60.23 & \textbf{60.85} & 1.03 & 46.48 & \textbf{**49.14} & 5.72 & 47.58 & \textbf{**50.29} & 5.70 & 51.11 & \textbf{52.00} & 1.74 \\ 
 & C & 60.43 & \textbf{61.28} & 1.40 & 44.61 & \textbf{46.31} & 3.80 & 46.21 & \textbf{**48.78} & 5.58 & 51.72 & \textbf{53.25} & 2.95 \\ 
 & LSC & 62.22 & \textbf{*64.44} & 3.58 & 47.63 & \textbf{**51.12} & 7.32 & 48.24 & \textbf{*50.70} & 5.11 & 53.51 & \textbf{*56.04} & 4.74 \\ 
 \cmidrule{1-2}
 
 \multirow{4}{*}{\makecell[tl]{\atomic+\\\dataset-C}} & L & 54.36 & \textbf{54.80} & 0.81 & 46.25 & \textbf{**51.57} & 11.51 & 42.81 & \textbf{43.13} & 0.76 & 50.71 & \textbf{50.81} & 0.20 \\ 
 & S & 54.50 & \textbf{55.75} & 2.29 & 45.59 & \textbf{*48.11} & 5.53 & 45.40 & \textbf{**48.50} & 6.83 & 51.00 & \textbf{51.78} & 1.52 \\ 
 & C & 54.43 & \textbf{55.50} & 1.97 & 42.61 & \textbf{43.26} & 1.53 & 43.11 & \textbf{44.13} & 2.38 & 51.44 & \textbf{*53.46} & 3.93 \\ 
 & LSC & 55.68 & \textbf{*57.91} & 4.01 & 47.11 & \textbf{**51.25} & 8.80 & 45.75 & \textbf{**49.03} & 7.18 & 52.44 & \textbf{**55.68} & 6.18 \\ 
 \cmidrule{1-2}
 
 \multirow{4}{*}{\makecell[tl]{\atomic+\\\dataset}} & L & 56.63 & \textbf{57.11} & 0.85 & 50.39 & \textbf{**54.52} & 8.20 & 47.92 & \textbf{48.27} & 0.73 & 53.11 & \textbf{53.41} & 0.56 \\ 
 & S & 56.53 & \textbf{57.42} & 1.57 & 48.92 & \textbf{**52.07} & 6.44 & 49.21 & \textbf{**52.51} & 6.72 & 53.10 & \textbf{53.67} & 1.09 \\ 
 & C & 56.40 & \textbf{57.64} & 2.21 & 47.96 & \textbf{48.30} & 0.70 & 48.16 & \textbf{50.00} & 3.82 & 53.90 & \textbf{55.48} & 2.94 \\ 
 & LSC & 58.27 & \textbf{60.53} & 3.87 & 50.25 & \textbf{**54.27} & 8.02 & 49.85 & \textbf{**53.09} & 6.50 & 54.50 & \textbf{**57.66} & 5.79 \\ 
 \bottomrule
\end{tabular}

\caption{For generations of COMET models trained on different subsets of \atomic{} and \dataset{}, the \Patthree{} scores of the \textit{all} and \textit{valid} sets determined by \textbf{L}, \textbf{S}, \textbf{C} and \textbf{LSC} discriminators with respect to the original and negation evaluation sets. The single (*) and double asterisks (**) indicate significance at p<0.05 and p<0.01 respectively. iprv\% is the percentage improvement of the \textit{valid} set over the \textit{all} set.}
\label{tab:full-p@3}
\end{table*}

\begin{table}[t!]
\small
\centering
\begin{tabular}{l|l|lc}

\toprule
\textbf{Event + Rel} & \textbf{Generation} & \textbf{V} & \textbf{P}\\
\midrule
 
 \multirow{5}{*}{\makecell[tl]{X does \\not skate \\around\\\textbf{\textit{xAttr}}}} & athletic & \xmark & \xmark \\
 & careless & \xmark & \xmark \\
 & lazy & \cmark & \cmark \\
 & uncoordinated & \cmark & \cmark \\
 & unskilled & \cmark & \cmark \\ 
 \midrule
 
 \multirow{5}{*}{\makecell[tl]{X does \\not sit \\behind Y\\\textbf{\textit{xIntent}}}} & to be alone & \cmark & \cmark \\
 & to be left alone & \cmark & \cmark \\
 & to avoid Y & \cmark & \cmark \\
 & to sit & \xmark & \xmark \\
 & to wait & \cmark & \xmark \\
 \midrule

 \multirow{5}{*}{\makecell[tl]{X does \\not look \\angry\\\textbf{\textit{xNeed}}}} & to calm down & \xmark & \cmark \\
 & to watch a movie & \cmark & \xmark \\
 & to have been provoked & \xmark & \xmark \\
 & to not be angry & \cmark & \cmark \\
 & to be calm & \cmark & \cmark \\
 \midrule
 
 \multirow{5}{*}{\makecell[tl]{X does \\not rent \\an apartment\\\textbf{\textit{xWant}}}} & to save money & \cmark & \cmark \\
 & to get a job & \cmark & \xmark \\
 & to pay rent & \xmark & \xmark \\
 & to move in & \xmark & \xmark \\
 & to get a new apartment & \cmark & \cmark \\
 \midrule
 
 \multirow{5}{*}{\makecell[tl]{X is \\not offered \\the job\\\textbf{\textit{oWant}}}} & to hire X & \xmark & \xmark \\
 & to fire X & \cmark & \cmark \\
 & to hire someone else & \cmark & \cmark \\
 & to accept the job & \xmark & \xmark \\
 & to hire them & \xmark & \xmark \\
 \midrule
 
 \multirow{5}{*}{\makecell[tl]{X does\\not buy \\a snack\\\textbf{\textit{xReact}}}} & satisfied & \xmark & \xmark\\
 & hungry & \cmark & \cmark\\
 & satiated & \xmark & \xmark\\
 & full & \xmark & \xmark\\
 & guilty & \cmark & \xmark\\
 \midrule
 
 \multirow{5}{*}{\makecell[tl]{X does \\not bring Y\\into conflict\\\textbf{\textit{oReact}}}} & relieved & \cmark & \cmark \\
 & sad & \xmark & \xmark \\
 & satisfied & \cmark & \cmark \\
 & grateful & \cmark & \cmark \\
 & angry & \xmark & \xmark \\
 \midrule
 
 \multirow{5}{*}{\makecell[tl]{X does \\not learn \\new things\\\textbf{\textit{xEffect}}}} & gains knowledge & \xmark & \xmark \\
 & becomes lazy & \cmark & \cmark \\
 & gets bored & \cmark & \cmark \\
 & becomes ignorant & \xmark & \cmark \\
 & cries & \cmark & \cmark \\
 \midrule
 
 \multirow{5}{*}{\makecell[tl]{X does \\not put Y \\in mind\\\textbf{\textit{oEffect}}}} & becomes confused & \xmark & \xmark \\
 & does not think about X & \cmark & \cmark \\
 & Y thinks about X & \xmark & \xmark \\
 & Y is not remembered & \cmark & \cmark \\
 & cries & \xmark & \cmark \\
\bottomrule

\end{tabular}
\caption{Randomly selected generations of the original COMET model regarding logical negation events in \dataset{}-L. The top 5 options are classified as either \textit{valid} or \textit{invalid} by the \textbf{LSC} discriminator. \textbf{V} indicates whether an option is classified as \textit{valid} by the \textbf{LSC} discriminator. \textbf{P} indicates whether an option is plausible judging by humans.} 
\label{tab:discriminator-generations-logical}
\end{table}

\begin{table}[t!]
\small
\centering
\begin{tabular}{l|l|lc}

\toprule
\textbf{Event + Rel} & \textbf{Generation} & \textbf{V} & \textbf{P}\\
\midrule
 
 \multirow{5}{*}{\makecell[tl]{X hardly ever\\increases X's\\knowledge\\\textbf{\textit{xAttr}}}} & intelligent & \xmark & \xmark \\
 & determined & \xmark & \xmark \\
 & studious & \xmark & \xmark \\
 & lazy & \cmark & \cmark \\
 & dedicated & \xmark & \xmark \\ 
 \midrule

 \multirow{5}{*}{\makecell[tl]{X avoids \\skating \\around\\\textbf{\textit{xIntent}}}} & to have fun & \xmark & \xmark \\
 & to be safe & \cmark & \cmark \\
 & to stay home & \cmark & \cmark \\
 & to stay in shape & \xmark & \xmark \\
 & to get fit & \xmark & \xmark \\
 \midrule
 
 \multirow{5}{*}{\makecell[tl]{X not at all \\wants to learn \\karate\\\textbf{\textit{xNeed}}}} & learn karate & \xmark & \xmark \\
 & to not like it & \cmark & \cmark \\
 & to avoid it & \cmark & \cmark \\
 & to be lazy & \cmark & \cmark \\
 & to find a teacher & \xmark & \xmark \\
 \midrule
 
 \multirow{5}{*}{\makecell[tl]{X refuses \\to hear a \\scary noise\\\textbf{\textit{xWant}}}} & to run away & \xmark & \xmark \\
 & to go to sleep & \cmark & \cmark \\
 & to be safe & \cmark & \cmark \\
 & to keep quiet & \cmark & \cmark \\
 & to avoid the noise & \cmark & \cmark \\
 \midrule
 
 \multirow{5}{*}{\makecell[tl]{X never \\brings Y into \\conflicts\\\textbf{\textit{oWant}}}} & to avoid X & \xmark & \xmark \\
 & to be left alone & \xmark & \cmark \\
 & to thank X & \cmark & \cmark \\
 & to fight back & \xmark & \xmark \\
 & to avoid conflict & \xmark & \cmark \\
 \midrule
 
 \multirow{5}{*}{\makecell[tl]{X scarcely\\gets sunburned\\ \\\textbf{\textit{xReact}}}} & burned & \xmark & \xmark\\
 & hurt & \xmark & \xmark\\
 & sick & \xmark & \xmark\\
 & sad & \xmark & \xmark\\
 & satisfied & \cmark & \cmark\\
 \midrule

 \multirow{5}{*}{\makecell[tl]{X under no\\ circumstances\\forgets Y's wallet\\\textbf{\textit{oReact}}}} & upset & \xmark & \xmark \\
 & sad & \xmark & \xmark \\
 & angry & \xmark & \xmark \\
 & thankful & \cmark & \cmark \\
 & grateful & \cmark & \cmark \\
 \midrule
 
 \multirow{5}{*}{\makecell[tl]{X has trouble \\with advertising \\X's business\\\textbf{\textit{xEffect}}}} & loses money & \cmark & \cmark \\
 & loses clients & \cmark & \cmark \\
 & gets fired & \cmark & \cmark \\
 & gets sued & \xmark & \xmark \\
 & cries & \cmark & \cmark \\
 \midrule

 \multirow{5}{*}{\makecell[tl]{X fails to \\make it through \\the day\\\textbf{\textit{oEffect}}}} & loses a friend & \cmark & \cmark \\
 & worries about X & \cmark & \cmark \\
 & worried & \cmark & \cmark \\
 & want them to do better & \cmark & \cmark \\
 & cries & \cmark & \xmark \\
\bottomrule

\end{tabular}
\caption{Randomly selected generations of the original COMET model regarding semi-logical negation events from \dataset{}-S. The top 5 options are classified as either \textit{valid} or \textit{invalid} by the \textbf{LSC} discriminator. \textbf{V} indicates whether an option is classified as \textit{valid} by the \textbf{LSC} discriminator. \textbf{P} indicates whether an option is plausible judging by humans.} 
\label{tab:discriminator-generations-semi-logical}
\end{table}


\begin{table}[t!]
\small
\centering
\begin{tabular}{l|l|lc}

\toprule
\textbf{Event + Rel} & \textbf{Generation} & \textbf{V} & \textbf{P}\\
\midrule
 
 \multirow{5}{*}{\makecell[tl]{X keeps \\the old one \\ \\\textbf{\textit{xAttr}}}} & determined & \xmark & \xmark \\
 & careful & \xmark & \xmark \\
 & loyal & \cmark & \cmark \\
 & tired & \xmark & \xmark \\
 & caring & \xmark & \xmark \\ 
 \midrule
 
 \multirow{5}{*}{\makecell[tl]{X pays twice \\the retail price \\for a rare item\\\textbf{\textit{xIntent}}}} & to own something rare & \cmark & \cmark \\
 & to buy something rare & \cmark & \cmark \\
 & to purchase something rare & \cmark & \cmark \\
 & to obtain something & \xmark & \xmark \\
 & to acquire something & \xmark & \xmark \\
 \midrule

 \multirow{5}{*}{\makecell[tl]{X is a \\complete  \\loner\\\textbf{\textit{xNeed}}}} & to avoid people & \cmark & \cmark \\
 & to be lonely & \cmark & \cmark \\
 & to start a relationship & \xmark & \xmark \\
 & to stay quiet & \cmark & \cmark \\
 & to get into trouble & \xmark & \xmark \\
 \midrule

 \multirow{5}{*}{\makecell[tl]{X is \\miserable  \\playing \\\textbf{\textit{xWant}}}} & to get better & \cmark & \cmark \\
 & to take a break & \xmark & \cmark \\
 & to go home & \xmark & \xmark \\
 & to go to sleep & \xmark & \xmark \\
 & to cry & \cmark & \cmark \\
 \midrule
 
 \multirow{5}{*}{\makecell[tl]{X is picky \\about \\everything\\\textbf{\textit{oWant}}}} & to ignore X & \cmark & \cmark \\
 & to avoid X & \cmark & \cmark \\
 & to talk to X & \xmark & \xmark \\
 & to help X & \xmark & \xmark \\
 & to make X feel better & \cmark & \xmark \\
 \midrule

 \multirow{5}{*}{\makecell[tl]{X resigns\\himself \\ \\\textbf{\textit{xReact}}}} & relieved & \cmark & \cmark\\
 & relaxed & \cmark & \xmark\\
 & satisfied & \xmark & \xmark\\
 & accomplished & \xmark & \xmark\\
 & sad & \cmark & \cmark\\
 \midrule

 \multirow{5}{*}{\makecell[tl]{X gives away \\X's laptop\\ \\\textbf{\textit{oReact}}}} & grateful & \cmark & \cmark \\
 & thankful & \cmark & \cmark \\
 & upset & \xmark & \xmark \\
 & sad & \xmark & \xmark \\
 & surprised & \cmark & \cmark \\
 \midrule

 \multirow{5}{*}{\makecell[tl]{X goes \\home \\ \\\textbf{\textit{xEffect}}}} & relaxes & \cmark & \cmark \\
 & goes to sleep & \cmark & \cmark \\
 & is greeted by family & \xmark & \cmark \\
 & gets rest & \cmark & \cmark \\
 & gets tired & \xmark & \xmark \\
 \midrule
 
 \multirow{5}{*}{\makecell[tl]{X puts Y \\out of mind \\ \\\textbf{\textit{oEffect}}}} & has a better day & \xmark & \xmark \\
 & becomes sad & \cmark & \cmark \\
 & cries & \cmark & \cmark \\
 & becomes grateful towards X & \xmark & \xmark \\
 & feels better & \xmark & \xmark \\
\bottomrule

\end{tabular}
\caption{Randomly selected generations of the original COMET model regarding commonsense contradiction events from \dataset{}-C. The top 5 options are classified as either \textit{valid} or \textit{invalid} by the \textbf{LSC} discriminator. \textbf{V} indicates whether an option is classified as \textit{valid} by the \textbf{LSC} discriminator. \textbf{P} indicates whether an option is plausible judging by humans.} 
\label{tab:discriminator-generations-pragmatic}
\end{table}

\subsection{Statistical Significance Testing} 
\label{app:permutation}
To compare \patlenvalid{} for the \textit{all} and \textit{valid} sets, we use a Permutation Test\footnote{http://rasbt.github.io/mlxtend/} with 1,000 permutations to test for statistical significance. For multiple comparisons, we use the Bonferroni method \cite{Haynes2013} to correct significance thresholds.

\subsection{Quality Check for the Human Evaluation}

\begin{figure*}[t]
    \centering
    \includegraphics[width=1\textwidth]{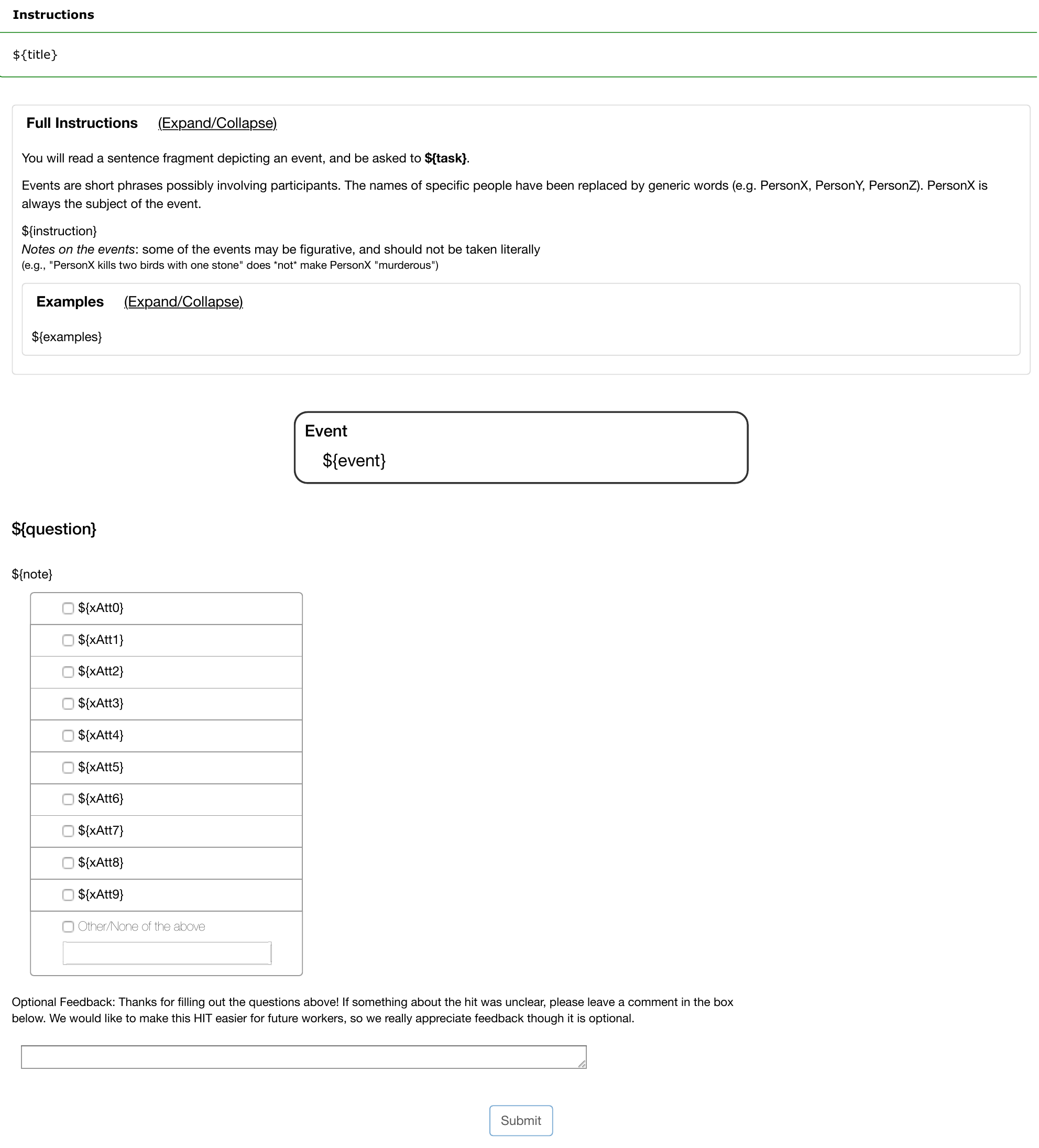}
    \caption{Snippet of the human evaluation task used to evaluate model generated tail inferences.}
    \label{fig:human_eval_mturk_hit}
\end{figure*}

We conduct comprehensive pre- and post-evaluation screening on the users and the tasks being completed to ensure the objectivity and high quality of the evaluations. Besides qualifying users during pilot batches, we double check to remove evaluation tasks that are not carefully conducted (\eg, tasks done by users that select all/no options for all hundreds of tasks that they perform). Figure \ref{fig:human_eval_mturk_hit} shows a snippet of the human evaluation MTurk task.


